\documentclass{article}

\usepackage[accepted]{icml2021}

\usepackage[]{natbib}

\usepackage[utf8]{inputenc} 
\usepackage[T1]{fontenc}    
\usepackage{hyperref}       
\usepackage{url}            
\usepackage{booktabs}       
\usepackage{amsfonts}       
\usepackage{nicefrac}       
\usepackage{microtype}      
\usepackage{xcolor}         
\usepackage{amsmath}
\usepackage{amssymb}
\usepackage{layouts}
\usepackage{graphicx}

\begin{document}

\twocolumn[
\icmltitle{Explaining Reinforcement Learning Policies through Counterfactual Trajectories}


\icmlsetsymbol{equal}{*}

\begin{icmlauthorlist}
\icmlauthor{Julius Frost}{bos}
\icmlauthor{Olivia Watkins}{ucb}
\icmlauthor{Eric Weiner}{hmc}
\icmlauthor{Pieter Abbeel}{ucb}
\icmlauthor{Trevor Darrell}{ucb}
\icmlauthor{Bryan Plummer}{bos}
\icmlauthor{Kate Saenko}{bos}
\end{icmlauthorlist}

\icmlaffiliation{bos}{Boston University}
\icmlaffiliation{ucb}{UC Berkeley}
\icmlaffiliation{hmc}{Harvey Mudd College}


\icmlcorrespondingauthor{Julius Frost}{juliusf@bu.edu}
\icmlcorrespondingauthor{Olivia Watkins}{oliviawatkins@berkeley.edu}

\icmlkeywords{Explainable AI, Reinforcement Learning}

\vskip 0.3in
]

\printAffiliationsAndNotice{}


\newcommand{\cn}{{\color{red}(citation needed)}}
\newcommand{\cc}[1]{{\color{red} #1}}
\newcommand{\ow}[1]{{{\color{purple}Olivia: #1}}}
\newcommand{\jf}[1]{{{\color{blue}Julius: #1}}}
\newcommand{\ks}[1]{{{\color{green}K: #1}}}

\begin{abstract}
In order for humans to confidently decide where to employ RL agents for real-world tasks, a human developer must validate that the agent will perform well at test-time. Some policy interpretability methods facilitate this by capturing the policy's decision making in a set of agent rollouts. However, even the most informative trajectories of training time behavior may give little insight into the agent's behavior out of distribution. In contrast, our method conveys how the agent performs under distribution shifts by showing the agent's behavior across a wider trajectory distribution. We generate these trajectories by guiding the agent to more diverse unseen states and showing the agent's behavior there. In a user study, we demonstrate that our method enables users to score better than baseline methods on one of two agent validation tasks.



\end{abstract}
\section{Introduction}



In most of the benchmarks where reinforcement learning (RL) has shown impressive results \cite{bellemare13arcade}, \cite{deepmindcontrolsuite2018}, agents are trained and tested in the same environment. 
In contrast, many practical applications involve sim-to-real transfer or a real-world train/test distribution shift which can hurt test-time performance \cite{eysenbach2020off}, \cite{akkaya2019solving}, \cite{rusu2017sim}, \cite{peng2018sim}.
Therefore it important to validate the performance and behavior of RL agents in the presence of a distribution shift.
To do this, we take a step beyond summary metrics and focus on explainability techniques which help humans gain a more in-depth understanding of agent behavior.

\begin{figure}
    \centering
    \includegraphics[width=.85\columnwidth]{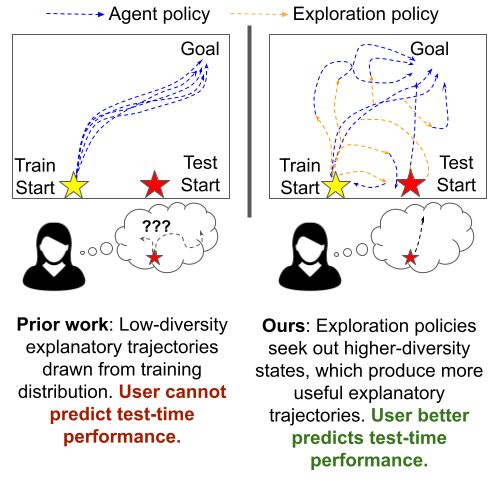}
    \caption{Our method seeks out more diverse states from which to show rollouts of agent behavior to the user.}
    \label{fig:teaser}
\end{figure}

Previous work in explainability often aims to help users understand and predict model behavior by explaining a set of model predictions \cite{hoffman2018metrics}. 
In the reinforcement learning setting, it may be desirable to select the most informative segments that show the behavior of the agent. 
With this motivation, critical states methods show the agent's behavior in states which the agent thinks are important \cite{huang_critical_states}, \cite{amir2018highlights}. 
However, even if these methods successfully explain policy behavior, we cannot conclude if the behavior generalizes when the distribution shifts in the environment.

Our work builds upon the core idea of creating a set of informative trajectories of the agent's behavior, but we focus on showing trajectories which are informative of the agent's behavior under a test-time state distribution. To achieve this, we first define a prior over the types of states we expect to see at test time. Unfortunately, most simple priors (e.g. uniform) will also include many unreachable states which are not informative to show the user. To avoid this issue, we use an exploration policy to seek out states which match our prior distribution. By navigating to these new states rather than directly initializing our environment in these states, we reach a diverse yet still reachable set of states. 

To summarize our contributions, we designed counterfactual trajectories that explain behavior policies in out-of-distribution states and conducted studies with human users that show increased understanding and generalization to out-of-distribution states by measuring prediction ability. \footnote{Our code is available at \url{https://github.com/juliusfrost/cfrl-rllib}}

\begin{figure*}
    \centering
    \includegraphics[width=0.75\textwidth]{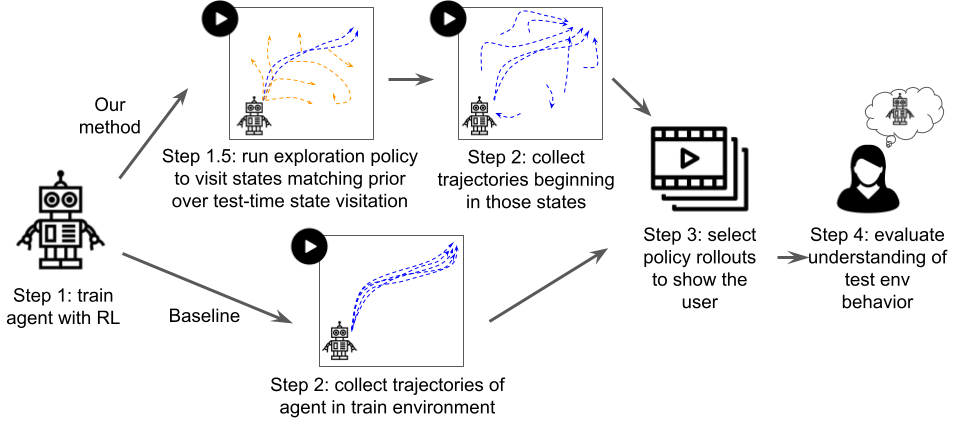}
    \caption{Explanation pipeline}
    \label{fig:method}
\end{figure*}

\section{Related Work}

\subsection{Saliency}
    One popular class of agent interpretability approaches is saliency methods, which aim to show which features from the input cause specific agent output. This method was employed by  \cite{greydanus2017visualizing},  \cite{hilton2020understanding}, and \cite{puri2019explain} to interpret which parts of the input that were deemed important to the agent's decision.  \cite{anderson2019explaining} use saliency as a part of a more general agent explanation method.
    Even with all of the success, saliency methods have been shown to have limitations. \cite{atrey2019exploratory} show that saliency does not necessarily correspond to underlying representation in RL. In addition, saliency methods are not designed to deal with agents with memory, and in general may not be informative in a multi-timestep environment. All of these challenges point to saliency as a helpful explanatory tool but not as a stand-alone solution. 
    
\subsection{Interpretable Representations}
    Another approach is to use interpretable intermediate representations. In this approach, models are constructed so that even if they are not interpretable end-to-end, there will be some interpretable bottleneck or set of features used to make the final decision.  \cite{madumal2019explainable} use a causal model to generate explanations. \cite{chen2015deepdriving} apply this method to autonomous driving, where a model is trained to take an image and output understandable intermediate representations in addition to a final action. Other work such as \cite{kim2018textual} and \cite{jiang2019language} use language as an informative intermediate representation or output. While successful in some applications, these kinds of interpretability methods often rely on domain knowledge to choose appropriate intermediates, and they are not applicable if a policy's decision cannot be explained by a small set of interpretable components.

\subsection{Critical States}
    In the critical states framework, explanations are generated by showing especially informative ``critical'' states or trajectories from an agent. For example the work of \cite{huang_critical_states} considers states critical if they have a large Q-value-difference between actions or a very low entropy action distribution under a maximum entropy learning regimen. \cite{amir2018highlights} employ a similar approach, but they use videos instead of states and filter for diversity of trajectories. These methods can be seen as extensions over a baseline method of visualizing random states or failure states of an agent in an environment. The explanation format of our method - videos of the agent's behavior in informative situations - is closest to this line of work.
    
\subsection{Counterfactual Style Explanations}
    In \cite{rupprecht2019finding}, researchers build a generative model of states and try to show users the agent's behavior in states that optimize some notion of "interesting." Again using a generative model, \cite{olson2019counterfactual} use a GAN to produce modifications to states such that in the new state an agent takes a different action than it would have in the original state. While these works synthesize states directly, our method uses valid actions to access different parts of the state space. This means that any state we visit is actually reachable, at least in the training environment. The closest method to ours appears in the work of \cite{witty2018measuring}, in which the authors try to characterize generalization by exploring different starting configurations, either by directly changing the start state or by using states visited by another agent. In contrast to our work, their insights are applied to characterizing generalization, and not as explanations for humans to understand agent behavior. 

\cite{anderson2019explaining} provide a detailed empirical user study on the effectiveness of saliency map and reward explanations. They conclude there is no one explanation that fits all instances, but using several methods yields the best mental model, and thus encourage using multiple methods for explainability. We hope to introduce a novel explainability method which may add to this suite of tools.

\section{Method}

\subsection{Preliminaries}
In this work, we consider the challenge of generating informative trajectories which help a human user understand an agent's policy (called the $\textbf{behavioral policy}$) $\pi_\theta (a_t | s_t)$, which is parametrized by weights $\theta$ which can be trained through any RL or imitation learning method. The agent takes actions $a_t$ in states $s_t$, resulting in a trajectory $\tau$.

\begin{figure}
    \centering
    \includegraphics[width=\columnwidth]{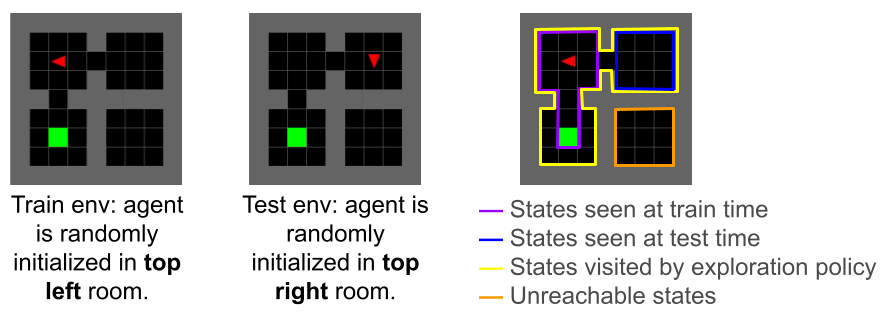}
    \caption{Distribution shifts, illustrated in the MiniGrid environment. Our user study was performed in an environment similar to this, except with the bottom-right room connected to the others.}
    \label{fig:dist_shift}
\end{figure}

Our goal is to help the user predict the agent's behavior in a new test environment in the presence of a state distribution shift which results in different trajectory probabilities at train and test time: $p^{\text{train}}(\tau | \theta) \neq p^{\text{test}}(\tau | \theta)$.  We consider two types of environment changes which can produce this distribution shift: differences in the initial state distribution ($p_0^{\text{train}}(s_0) \neq p_0^{\text{test}}(s_0)$ for some initial state $s_0$) and slight differences in the dynamics ($p^{\text{train}}(s_{t} | s_{t-1}, a_{t-1}) \neq p^{\text{test}}(s_{t} | s_{t-1}, a_{t-1})$ for some state $s_t$ and action $a_t$). We assume that there are no changes to the agent's action space or policy at test time  - i.e. $\pi_\theta(a_{t} | s_{t})$ remains consistent between train and test time. We also assume that dynamics distribution shifts are minor. See Section \ref{sec:where_we_help} for more detailed discussion on this.

\subsection{Method Overview}

Figure \ref{fig:method} summarizes our method. We build upon the explanatory pipeline used in past critical states work such as \cite{huang_critical_states}. An agent is first pretrained using any RL or imitation learning algorithm. We then collect many trajectory rollouts of the agent, select a set to show the user, and finally test whether these explanatory trajectories have helped the human better predict the agent's behavior in the test environment.  This is challenging because of the distribution shift between the train and test environments.

Past work has attempted to select an informative set of trajectories to show the user by innovating on Step 3 in Figure \ref{fig:method} - the trajectory selection process. The critical states line of work selects trajectory segments which show the agent's behavior at states where the policy believes one particular action is much better than the others \cite{huang_critical_states}. Unfortunately, methods like these which sub-select trajectories from a set of agent rollouts in the train environment are often not very informative about the agent's behavior under a distribution shift. Consider the train/test initial state distribution shift in Figure \ref{fig:dist_shift}. A well-trained agent which was always initialized in the top-left room during training may take a nearly-deterministic trajectory navigating toward the goal every time, so none of the trajectories visited by the agent in the training environment will ever visit the states in the top-right room which the agent will experience at test time.

One way to generate trajectories in the training distribution which are more informative of the test distribution would be to manually change the starting state distribution.  For instance, in the grid world shown in Figure \ref{fig:dist_shift}, the simulator could be modified to directly initialize the agent in a uniform distribution over starting cells, ensuring that some trajectories will have been seen in whatever state the agent finds itself at test time.

One challenge with this approach is that it is often difficult to restrict the set of randomized initialization to reachable states the agent could possibly visit at test time. Instead, we propose creating a new distribution of start states using an \textbf{exploratory policy} $\pi_{\phi}$ which can navigate from the states seen on the training distribution to a new, more diverse distribution of start states $p_0^{\text{expl}}(s_0 | \phi)$ which are guaranteed to be reachable. Implementing this only involves adding one additional Step 1.5 to the pipeline shown in Figure \ref{fig:method}.

Our trajectory generation method is summarized in the following algorithm:
\begin{enumerate}
    \item Run the behavioral policy $\pi_{\theta}$ in the train environment. Pause at a randomly selected state along the trajectory.
    \item Run exploratory policy $\pi_{\phi}$ starting at this state for a fixed number of timesteps. The exploratory policy will guide the agent to a set of states not typically observed in the training data. We call these \textbf{counterfactual states} because we can frame it as a counterfactual question: ``what would the behavioral policy do if it instead navigated to this state?''
    \item Run the behavioral policy starting from the counterfactual state for the remainder of the trajectory.
    \item Show the user video rollouts of the agent's behavior starting in these counterfactual states.
\end{enumerate}

\begin{figure}
    \centering
    \includegraphics[width=\columnwidth]{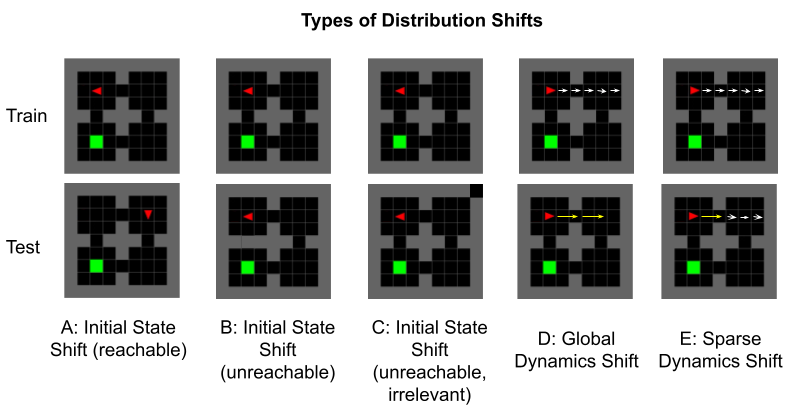}
    \caption{Possible types of train/test distribution shifts. Our user experiments use Shift A, although we would expect our method to be applicable to Shifts C and E as well.}
    \label{fig:dist_shift_compared}
\end{figure}

Note that the trajectories selection method used in item 4 of our algorithm can be very simple - in our experiments we randomly select trajectories. However, our method is complementary to other work on selecting informative trajectories, so future work could select trajectories more systematically, for instance filtering out near-repeated trajectories as was done in \cite{amir2018highlights} or selecting based on critical states as was done in \cite{huang_critical_states}.

\subsection{Exploration Objective}
\label{sec:expl_obj}
Our aim is to choose an exploration objective which will guide the agent to states which produce maximally informative trajectories. Intuitively, explanatory trajectories will be more informative the closer the distribution of trajectories visited by our exploratory policy is to the test-time distribution of trajectories visited. More formally, let $p_0^{\text{expl}}(s_0 | \phi)$ be the distribution of states in which the agent finds itself after running the exploration policy, and let $p^{\text{expl}}(\tau | \phi, \theta)$ be the distribution of trajectories generated by rolling out the behavioral policy $\pi_{\theta}ß$ initialized in $p_0^{\text{expl}}(s_0 | \phi)$. Our goal is to achieve $p^{\text{expl}}(\tau | \phi, \theta) \approx p^{\text{test}}( \tau | \theta)$.

It is easiest to choose a principled exploration objective by making the assumption that train and test-time dynamics are approximately identical. In this situation, the train-test distribution shift consists entirely of changes to the initial state: $p_0^{\text{train}}(s_0) \neq p_0^{\text{test}}(s_0)$. However, if we are able to choose an exploration policy which seeks out states matching the test time distribution - i.e. if $p_0^{\text{expl}}(s_0 | \phi) \approx p_0^{\text{test}}(s_0)$ - then because we roll out the same behavioral policy in environments with identical dynamics starting from nearly identical start distributions, we will achieve $p^{\text{expl}}(\tau | \phi, \theta) \approx p^{\text{test}}( \tau | \theta)$.


We do not precisely know $p_0^{\text{test}}(s_0)$ before the agent experiences the test environment, so instead we choose a prior for our test time start state distribution. In the absence of domain knowledge about what states are likely under the test-time distribution shift, we assume a uniform prior over reachable test-time states and try to match this distribution as closely as possible by using an exploratory policy which seeks out a uniform distribution of states. (Note that with this exploration objective, the only reason we begin to run $\pi_{\phi}$ along a trajectory generated by $\pi_{\theta}$ rather than than running $\pi_{\phi}$ from the start is to make achieving the exploration objective easier for a learned exploration policy.)  However, if the user does have domain knowledge about which states are likely at test time, the exploratory policy could be modified to preferentially seek out those states. For instance, if the user believes that test-time states will be near the training distribution, the exploration objective may instead be to seek out a uniform distribution of states within a certain radius of those seen at train time. 

There has been a variety of past exploration work in which agents are trained to seek out a uniform distribution of states or to learn distinct skills which take the agent to different parts of the state space such as \cite{sharma2019dynamics} and \cite{eysenbach2018diversity}. While any of these off-the-shelf exploration algorithms could be used, in our work, we experiment in the Minigrid environment, where the state space and dynamics are simple enough it is possible to hard-code an oracle exploration policy which achieves a uniform coverage over reachable states. By using this oracle policy, we are able to directly test the question of whether highly diverse states serve as better explanations rather than implicitly also testing the ability of the exploration policy.

\subsection{Distribution shifts where Our method helps}
\label{sec:where_we_help}
There are multiple types of changes between train and test environments which can induce a difference between $p^{\text{train}}(\tau | \theta)$ and $p^{\text{test}}(\tau | \theta)$, as illustrated in Figure \ref{fig:dist_shift_compared}. In Section \ref{sec:expl_obj}, we gave an intuitive justification for why we would expect our method to help in cases where train and test-time dynamics are identical, and where the exploration policy is able to cover the test-time start distribution - Shift A, in Figure \ref{fig:dist_shift_compared}. Now, we consider our method's usefulness in the presence of other distribution shifts.


Shift B starts the agent in a new state which is unreachable in the train environment, so it is unlikely that our explanations can help there. This restriction to only reachable distribution shifts is very limiting, but we can partially relax it in cases where the distribution shift does not change the agent's internal representation of the environment. Shift C illustrates an example of this: the test-time environment has the top-right corner wall segment removed, but so long as the agent's internal representation of the environment does not change much based on this change, the agent will act the same way it would act if the wall was present. As a result, explanations generated in the train environment will still be informative about test-env behavior.

Finally, we consider dynamics shifts. In the case of global dynamics shifts, as shown in Shift D, our method is likely unhelpful, since train-time and test-time trajectories look very different even when starting from the same state. However, if dynamics shifts only occur in a few states in the test environment as shown in Shift E, then most of the time a train-time trajectory and a test-time trajectory beginning at the same state will look identical. Therefore, we can think of the dynamics shift purely as a way to introduce a state distribution shift. While our explanations will not help the user predict these dynamics shifts, the user should be able to predict how the agent will act after the dynamics shift leads the agent to an unseen state.




\section{Experiments}
In this section, we test our primary hypothesis that diverse states help the user understand the behavioral policy's performance in the presence of a distribution shift. We 
empirically evaluate this through a Mechanical Turk user study for which we obtained IRB approval.
\subsection{Study Design}

\begin{figure}
    \centering
    \includegraphics[width=\columnwidth]{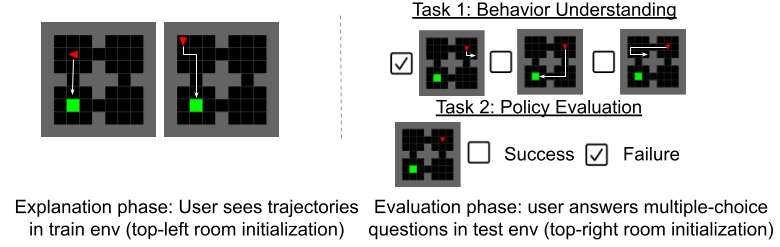}
    \caption{Experimental setup for the counterfactual states user study. Left: users see 10 train-time rollouts such as these in the explanation phase; Right: users are asked to either select the test-time behavior which corresponds with the policy they saw previously (Task 1 - Behavior Understanding) or predict the agent's success in a new state (Task 2 - Policy Evaluation).}
    \label{fig:cf_experiments}
\end{figure}

Our study consisted of a questionnaire hosted online. Our participants were Mechanical Turkers from the United States with a high Mechanical Turk approval rating on past jobs. Each questionnaire consisted of an explanation phase in which the user sees video trajectories of the policy in the train environment and an evaluation phase which measures the user's understanding of the policy's performance in the test environment. Train and test environments differ in which room the agent is initialized, as shown in Figure \ref{fig:cf_experiments}. We tested participants on two tasks, described in Sections \ref{sec:behavior_understanding_task} and \ref{sec:peformance_evaluation_task}.


\subsubsection{Explanation Types}
\label{sssec:explanation_types}
For each of the tasks, our user study compares three different types of explanations. In the \emph{Random States} setting, participants are shown 10 sample trajectories selected at random from a dataset of rollouts in an environment with the agent initialized in a particular start region. In the \emph{Critical States} setting, participants are shown a trajectory containing each of the 10 lowest entropy states in the dataset, all with the agent initialized in the same starting region each time. If two or more of the low-entropy states are in the same trajectory, the next lowest entropy states are selected. Finally, in the \emph{Counterfactual States} setting, participants are shown 10 trajectories of the behavioral policy beginning from a state chosen by the oracle exploration policy, which seeks out a uniform distribution over reachable states. The \emph{Random States} setting acts as a baseline for the other two methods, because it is common practice to look at random rollouts to understand policy behavior. The \emph{Critical States} setting was chosen as another baseline, because it is the prior work with an explanation format which most closely matches our explanations, allowing for easier comparison. Finally, the \emph{Counterfactual States} condition is our proposed explanation method. Since the agent is initialized in a different distribution of start states at test time, we expect the \emph{Random States} and \emph{Critical States} methods to be uninformative at helping the user predict test-time performance. In contrast, \emph{Counterfactual States} show more diversity and should enable to the user to predict test-time performance.

\subsubsection{Behavior Understanding Task}
\label{sec:behavior_understanding_task}

In this task, users first see videos of rollouts of $\pi_{\theta}$ in the training environment selected using one of the three explanation methods described in section \ref{sssec:explanation_types}. In this task, as users watch these explanations, they are asked to build a mental model of the specific behaviors of $\pi_{\theta}$. They are then presented with a context state in the test environment, which is identical to the train environment except that the agent is initialized in a different part of the state space where $\pi_{\theta}$ performs poorly.  Below the context video, the user is presented with videos of three potential continuation trajectories from the context state. Only one of the continuation trajectories is generated by $\pi_{\theta}$, and participants must select the continuation corresponding to $\pi_{\theta}$. The incorrect choices for continuation trajectories were generated with policies manually designed to have behavior which is distinctly different from $\pi_{\theta}$ - one of which is always successful in the test distribution, and one which fails in a way which is visually distinct from $\pi_{
\theta}$.

%
\subsubsection{Performance Evaluation Task}
\label{sec:peformance_evaluation_task}
We next evaluate whether the user can predict the performance of $\pi_{\theta}$ in the test environment.
To do this, users are shown a context state from a test environment where the agent starts in a different state. The user must guess whether $\pi_{\theta}$ will succeed or fail from that state, where success means the agent ends in the desired goal location. This task is designed to measure the participants' understanding of the probability an agent succeeds in its task given its state. 
\subsubsection{Experiment Settings}
We test our method in a custom Minigrid environment \cite{gym_minigrid} where there are four rooms with doors connecting between them (shown in Figure \ref{fig:dist_shift_compared}A). 
We train our behavioral policy using the A2C algorithm \cite{mnih2016asynchronous} to good performance on the train environment.
To select counterfactual states, we use an oracle policy which uniformly samples a feasible state and navigates towards it.
In the study, each participant sees 10 explanations and then answers 10 questions of our chosen evaluation task. 

For the Behavior Understanding task, the policy was trained in the top-left room, and explanations were generated in this room. In the test environment, the agent is positioned in the bottom-right room. As a result, the agent typically succeeds when initialized in the top-left (where we collect explanations) but typically fails when initialized in the top-right room (where we initialize in the evaluation phase).

For the Policy Evaluation task, we test users on two different behavioral policies. The first is only able to succeed starting in the top-left room, and the other is only able to succeed from the bottom-right room. In both cases, we collect explanations in the top-left room. Users are evaluated on the agent's behavior in all four rooms. 

%
%
\begin{table}
  \caption{Minigrid Policy Understanding and Evaluation}
  \label{table:results}
  \centering
\begin{tabular}{lllll}
\toprule
& Task 1 && Task 2& \\
\cmidrule(r){2-3} \cmidrule(r){4-5}
Condition      & Accuracy        & n     & Accuracy & n           \\
\midrule
Random         & 0.04            & 10    & 0.6315   & 19           \\
Critical       & 0.1458          & 10    & 0.6411   & 17           \\
Counterfactual & \textbf{0.3638} & 10    & 0.6904   & 21           \\
\bottomrule
\end{tabular}
\caption{Task 1 is the Behavior Understanding task. Task 2 is Performance Evaluation. Bold means the result is statistically greater than the rest for $p<0.05$ with a one-sided T-test.}
\end{table}
\subsection{Results}
Qualitatively, in both tasks, the counterfactual states method produces more informative rollouts.  In both tasks, neither the random states nor critical states methods produce any explanatory trajectories where the agent visits the room where the it is initialized at test time. This makes it hard for the user to predict the agent's performance in these states. However, in the counterfactual states condition, the user sees multiple examples of the agent in the distribution of states it will see at test time.

Quantitative results, however, are mixed. Table \ref{table:results} shows that for the Behavior Understanding task, users are able to select the correct behavior $36.38\%$ of the time, a statistically significant improvement over the random and critical states conditions. This performance is only marginally above random guessing ($33.3\%$), but at least the user did not seem to be actively misled by the explanations, as occurred in the random and critical states cases. On the Policy Evaluation task, users perform similarly in all conditions, with no statistically significant difference between them. 

\section{Discussion}
 While our method outperformed baselines in one of two tasks, there is significant room for improvement, as users did not consistently perform well in any condition of any task, even though the environment and agent behavior were both quite simple. Part of the difficulty in constructing good explanations derives from the population we were testing. Mechanical Turkers likely have little prior experience with reinforcement learning or time to carefully analyze agent behavior. Anecdotally, we also observed that many survey-takers come into the task with their own biases and preconceptions (for instance, assuming that a policy will be deterministic rather than stochastic, or assuming that if an agent does well in one situation it probably does well everywhere). Future work could test the usefulness of our counterfactual states method for agent developers who can spend time to familiarize themselves with the task, the explanation interface, and the types of behaviors which typically emerge in policies. In this setting, we could also test our method's usefulness in more complex environments. The experiments we have run so far are quite limited: the environment is simple, an oracle exploration policy is available, the distribution shift only changes the agent’s starting position, and it is easy for a user to distinguish between correct and incorrect behaviors. Future work could explore whether counterfactual states hold value in more realistic settings.

In this paper, we hypothesized that showing higher-diversity reachable trajectories will enable users to better understand agent behavior. We tested this hypothesis through a user study in a gridworld in which the test-time initial state distribution differed from the train-time start distribution. Our method showed improvements over less diverse baseline methods in one of two tasks, but our experiments show room for improvement and illustrate the ongoing challenge of creating explanations which improve the understanding of non-expert users.
\section{Acknowledgements}
Thanks to Anna Rohrbach for ongoing feedback on this work.

{\small
\bibliographystyle{icml2021}
\bibliography{refs}

\appendix
}

\end{document}